\documentclass[conference,a4paper]{IEEEtran}
\IEEEoverridecommandlockouts

\usepackage[hidelinks]{hyperref}
\usepackage[cmex10]{amsmath}%American Math Society(AMS) math formatting
\usepackage{amssymb,amsfonts}%AMS extra symbols and fonts
\usepackage{dblfloatfix}%fix double column figure ordering and placement
\usepackage{multirow}
\usepackage[normalem]{ulem}
\useunder{\uline}{\ul}{}

\usepackage[ruled,vlined]{algorithm2e}
\usepackage{graphicx}
\graphicspath{{Figures/PDF/}{Figures/PNG/}}

\usepackage{booktabs}
\usepackage{siunitx}
\usepackage[numbers,compress]{natbib}
\usepackage{texnames}
\usepackage{bm,bbm}
\usepackage{orcidlink}
\usepackage{tikz}
\usetikzlibrary{arrows.meta, shapes.geometric, calc, positioning, fit}
\usetikzlibrary{arrows.meta,positioning,fit}
\begin{document}

% \title{\uppercase{FAConvLSTM: Factorized-Attention Convolutional LSTM for Efficient Feature Extraction in Multivariate Climate Data}
\title{{Factorized-Attention ConvLSTM for Efficient Feature Extraction in Multivariate Climate Data}
\thanks{This work is supported by NSF grants: CAREER: Big Data Climate Causality (OAC-1942714) and HDR Institute: HARP - Harnessing Data and Model Revolution in the Polar Regions (OAC-2118285)}
}

\author{\IEEEauthorblockN{Francis N.\ Nji\orcidlink{0009-0009-6559-4659}}
	\IEEEauthorblockA{\textit{University of Maryland, Baltimore County}\\
		Baltimore, USA\\
		fnji1@umbc.edu}
	% \and
	% \IEEEauthorblockN{Vandana Janeja\orcidlink{0000-0003-0130-6135}}
	% \IEEEauthorblockA{\textit{University of Maryland, Baltimore County}\\
	% 	Baltimore, USA\\
	% 	vjaneja@umbc.edu}
	\and
	\IEEEauthorblockN{Jianwu Wang\orcidlink{0000-0002-9933-1170}}
	\IEEEauthorblockA{\textit{University of Maryland, Baltimore County}\\
		Baltimore, USA\\
		jianwu@umbc.edu}
}

% \author{
% \IEEEauthorblockN{
% Francis N.~Nji\orcidlink{0009-0009-6559-4659},
% Vandana~Janeja\orcidlink{0000-0003-0130-6135},
% Jianwu~Wang\orcidlink{0000-0002-9933-1170}
% }
% \IEEEauthorblockA{
% \textit{University of Maryland, Baltimore County}\\
% Baltimore, USA\\
% fnji1@umbc.edu,\; vjaneja@umbc.edu,\; jianwu@umbc.edu
% }
% }

\maketitle

\begin{abstract}
% Learning physically meaningful spatial, temporal, and spatiotemporal representations from high-resolution multivariate Earth observation data remains a fundamental challenge due to strong local dynamics, teleconnection-scale dependencies, multi-scale variable interactions, and pronounced nonstationarity. Convolutional Long Short-Term Memory (ConvLSTM2D) networks are a widely used baseline for such tasks. However, their dense convolutional gating incurs prohibitive computational and memory costs at modern grid resolutions, and their strictly local receptive fields limit the ability to capture long-range spatial structure and disentangled latent dynamics critical for climate and remote sensing applications. 
Learning physically meaningful spatiotemporal representations from high-resolution multivariate Earth observation data is challenging due to strong local dynamics, long-range teleconnections, multi-scale interactions, and nonstationarity. While ConvLSTM2D is a commonly used baseline, its dense convolutional gating incurs high computational cost and its strictly local receptive fields limit the modeling of long-range spatial structure and disentangled climate dynamics. To address these limitations, we propose \emph{FAConvLSTM}, a Factorized-Attention ConvLSTM layer designed as a drop-in replacement for ConvLSTM2D that simultaneously improves efficiency, spatial expressiveness, and physical interpretability. FAConvLSTM factorizes recurrent gate computations using lightweight $1\times1$ bottlenecks and shared depthwise spatial mixing, substantially reducing channel complexity while preserving recurrent dynamics. Multi-scale dilated depthwise branches and squeeze-and-excitation recalibration enable efficient modeling of interacting physical processes across spatial scales, while peephole connections enhance temporal precision. To capture teleconnection-scale dependencies without incurring global attention cost, FAConvLSTM incorporates a lightweight axial spatial attention mechanism applied sparsely in time. A dedicated subspace head further produces compact per-timestep embeddings refined through temporal self-attention with fixed seasonal positional encoding. Experiments on multivariate spatiotemporal climate data shows superiority demonstrating that FAConvLSTM yields more stable, interpretable, and robust latent representations than standard ConvLSTM, while significantly reducing computational overhead.
\end{abstract}


\section{Introduction}
\label{sec:introduction}
Modern climate reanalyses and Earth observation products (e.g., ERA5 \cite{hersbach2020era5, nji2025b}, CARRA \cite{C3S_CARRA_2021}, satellite-derived geophysical fields \cite{schmugge2002remote, hersbach2020era5}) produce high-dimensional \emph{multivariate spatiotemporal tensors} that encode complex physical processes across space, time, and variables. These data exhibit several challenging characteristics: 
(i) \emph{strong local dynamics}, such as atmospheric fronts, storms, and melt onset events \cite{holton2012dynamic, markus2009recent, serreze2009atmospheric, wang2011snowmelt}; 
(ii) \emph{long-range spatial dependencies}, commonly referred to as teleconnections \cite{IPCCAR6WG1}, where distant regions interact through large-scale circulation patterns; 
(iii) \emph{multi-scale coupling} across physical variables and spatial resolutions \cite{Masson2025WRFCoupling, Li2025SoilMoistureCoupling}; and 
(iv) \emph{nonstationary temporal evolution} driven by seasonal forcing, episodic extreme events, and multi-scale climate variability \cite{guo2025data, ZantoutEtAl2025ShiftingDominantPeriods, JorgensenNielsenGammon2024NonstationarityReturnValues}. 
% Formally, we consider a multivariate spatiotemporal climate sequence to be: \(\mathbf{X} = \{\mathbf{X}_t\}_{t=1}^{T}, \quad \mathbf{X}_t \in \mathbb{R}^{H \times W \times C},\) where $T$ denotes the sequence length, $(H,W)$ the spatial grid resolution, and $C$ the number of physical variables (e.g., temperature, wind components, snow depth). 
A key objective is to learn \emph{physically latent} representations that (i) preserve local coherent structures (fronts, melt edges), (ii) capture broad-scale spatial modes (blocking patterns, teleconnections), and (iii) represent temporally persistent regimes and transitions: \(\mathbf{Z}_t = f_{\theta}(\mathbf{X}_{1:t}) \in \mathbb{R}^{H'\times W'\times D}, \quad D \ll C,\) such that $\mathbf{Z}_t$ supports reconstruction/forecasting and downstream analysis (e.g., clustering regimes, detecting rapid melt regions, or anomalous events).
% ConvLSTM networks~\cite{shi2015convlstm} extend classical LSTMs by replacing fully connected transformations with spatial convolutions, making them a de facto baseline for spatiotemporal modeling in climate and remote sensing applications. \(i_t = \sigma(W_{xi} * X_t + W_{hi} * H_{t-1} + b_i),\) and \(C_t = f_t \odot C_{t-1} + i_t \odot \tanh(W_{xc} * X_t + W_{hc} * H_{t-1}),\) where $*$ denotes spatial convolution and $\odot$ denotes element-wise multiplication. This design preserves spatial structure and enables limited spatiotemporal modeling.
ConvLSTM networks~\cite{shi2015convlstm} preserves spatial structure and enables limited spatiotemporal modeling. Despite their success, standard ConvLSTM2D layers suffer from four critical limitations and modeling constraints for complex multidimensional spatiotemporal climate data: 1) \textit{High compute/memory from per-gate dense convolutions:} Each gate uses separate $K{\times}K$ convolutions for input-to-state and state-to-state transitions. 
% With $F$ hidden channels, complexity scales as \(\mathcal{O}\!\left(T \cdot H W \cdot K^2 \cdot (C F + F^2)\right),\)
% multiplied by four gates, which becomes prohibitive for high-resolution grids and long sequences: 
2) \textit{Limited efficient long-range spatial interaction:} The convolutional receptive field grows slowly with depth/time.
% teleconnections require either very deep stacks or large kernels, both costly and prone to optimization instability. 
3) \textit{Channel entanglement and reduced physical interpretability:} Full convolutions mix channels densely at every gate, often learning representations that are difficult to align with physically meaningful modes (e.g., separable local advection vs.\ large-scale circulation). 4) \textit{Rigid grid inductive bias:} ConvLSTM assumes regular lattices; yet climate processes can be better modeled using graph relations (e.g., distance, orography, coastlines, flow connectivity), motivating attention on graph-structured spatial domains \cite{lam2023graphcast}.
% These limitations motivate the need for a more efficient and physically aligned recurrent architecture. 
To this end, we propose Factorized-Attention ConvLSTM (FAConvLSTM) that addresses the above limitations through principled architectural factorization and selective attention mechanisms. FAConvLSTM explicitly captures long-range spatial dependencies (teleconnections), statistical relationships between geographically distant climate regions through factorized attention and global context modeling. Our design is guided by four core principles: \textit{a) Factorization of spatial mixing and gating} to reduce computational complexity; \textit{b) Multi-scale spatial processing} to capture physical processes at different resolutions; \textit{c) Explicit temporal abstraction} via compact subspace embeddings and temporal attention; \textit{d) Efficient long-range spatial modeling} using lightweight axial attention applied sparsely in time. Our implementation code is publicly available
\footnote{https://github.com/big-data-lab-umbc/FAConvLSTM}.

\section{Related Work}
\label{sec:related}

% \textbf{ConvLSTM for spatiotemporal geoscience}.
ConvLSTM was introduced for precipitation nowcasting~\cite{shi2015convlstm} and now common for spatiotemporal remote sensing tasks. Recent studies have demonstrated its effectiveness for modeling spatiotemporal dependencies in geophysical and Earth observation applications. Leng et al. employed ConvLSTM to predict regional land subsidence by jointly capturing spatial correlations and temporal evolution from multi-temporal geospatial data, showing that ConvLSTM outperformed traditional machine learning baselines in long-term deformation forecasting~\cite{leng2023spatio}. Yao et al. integrated ConvLSTM with SBAS-InSAR time-series data to model surface deformation in mining areas, leveraging ConvLSTM’s convolutional gating mechanism to preserve spatial structure while learning temporal deformation dynamics~\cite{yao2023convlstm}. Chen et al. applied ConvLSTM to short-term seismic risk prediction induced by large-scale coal mining, formulating seismic intensity maps as spatiotemporal sequences and demonstrating that ConvLSTM effectively learns localized spatial interactions and temporal risk propagation~\cite{chen2023convlstm}.

\section{Problem Formulation}

Let us consider multivariate spatiotemporal 4-D tensor \(\mathbf{X} = \{\mathbf{X}_t\}_{t=1}^{T}, 
\quad \mathbf{X}_t \in \mathbb{R}^{H \times W \times C},\) where $T$ is the sequence length, $(H,W)$ the spatial grid resolution, and $C$ the number of physical variables.
Our objective is to learn latent representations that simultaneously capture:
(i) \emph{spatial dependencies} within each timestep: \(\mathbf{H}_t = f_{\text{spatial}}(\mathbf{X}_t; \theta_s),
\quad \mathbf{H}_t \in \mathbb{R}^{H \times W \times F},\) where $f_{\text{spatial}}(\cdot)$ is a learnable spatial encoder,
and $F$ is the number of latent spatial channels.
(ii) \emph{temporal dependencies} across timesteps: \((\mathbf{H}_t, \mathbf{C}_t)
=
f_{\text{temporal}}(\mathbf{H}_{t-1}, \mathbf{C}_{t-1}, \mathbf{X}_t; \theta_t),\) where $\mathbf{C}_t$ denotes the memory state, $f_{\text{temporal}}(\cdot)$ models temporal evolution via gated recurrence and/or attention-based mechanisms, and $\theta_t$ are temporal parameters, and (iii) \emph{spatiotemporal interactions} coupling space and time: \(\mathbf{H}_{1:T} = 
f_{\text{st}}(\mathbf{X}_{1:T}; \theta_{st}),\) where $f_{\text{st}}$ jointly models spatial mixing and temporal recurrence,
while remaining computationally efficient and physically interpretable.
\textit{Unified Learning Objective}: The overall learning problem is formulated as minimizing a composite objective: \(\min_{\theta}
\;
\mathcal{L}_{\text{task}}
+
\lambda_s \sum_{t=1}^{T} \mathcal{R}_{\text{spatial}}(\mathbf{H}_t)
+
\lambda_t \mathcal{R}_{\text{temporal}}(\mathbf{H}_{1:T}),\) where: $\theta = \{\theta_s, \theta_t, \theta_{st}\}$ denotes all learnable parameters and $\mathcal{L}_{\text{task}}$ denotes a downstream objective (e.g., reconstruction, forecasting, or clustering), $\mathcal{R}_{\text{spatial}}(\cdot)$ enforces spatial coherence and smoothness across the grid, and $\mathcal{R}_{\text{temporal}}(\cdot)$ regularizes temporal consistency or attention dynamics. The hyperparameters $\lambda_s$ and $\lambda_t$ control the relative strength of spatial and temporal regularization, respectively.

\section{Methodology}
\label{sec:methodology}

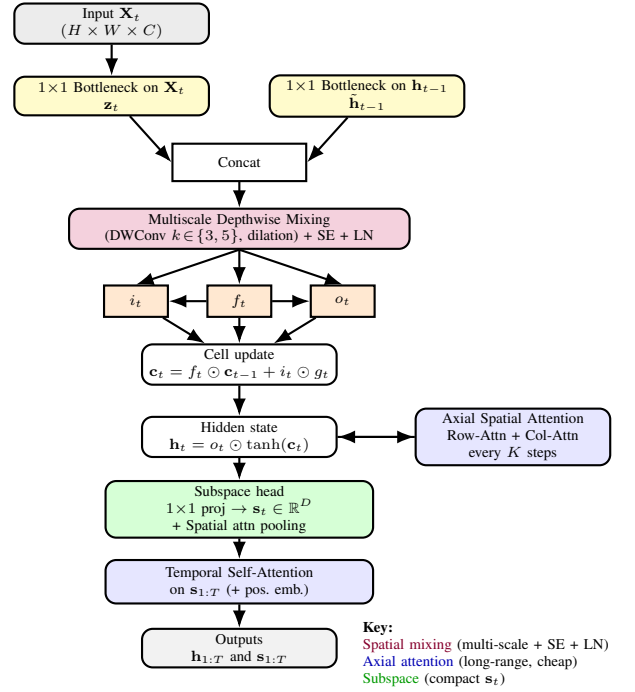
\begin{figure}[t]
\centering
\begin{tikzpicture}[
    x=1cm, y=1cm,
    font=\scriptsize,
    >=Latex,
    scale=0.80, transform shape,
    block/.style={draw, thick, rounded corners, align=center, minimum height=0.55cm, inner sep=2.2pt},
    op/.style={draw, thick, align=center, minimum height=0.62cm, inner sep=2.2pt},
    gate/.style={draw, thick, fill=orange!18, align=center, minimum height=0.52cm, minimum width=1.05cm, inner sep=1.4pt},
    attn/.style={draw, thick, rounded corners, fill=blue!10, align=center, minimum height=0.72cm, inner sep=2.2pt},
    subspace/.style={draw, thick, rounded corners, fill=green!16, align=center, minimum height=0.72cm, inner sep=2.2pt},
    arrow/.style={->, thick},
    flow/.style={->, thick, rounded corners=2pt},
    dasharrow/.style={->, thick, dashed, rounded corners=2pt},
    note/.style={align=left, font=\scriptsize}
]

% -------------------------
% Layout coordinates (manual)
% -------------------------
% Column width target ~ 8.5 cm => keep x-span ~ 8.0
% Top-down pipeline with a right-side axial refinement loop

% Input
\node[block, fill=gray!15, minimum width=3.2cm] (X) at (0, 6.2)
{Input $\mathbf{X}_t$\\$(H\times W\times C)$};

% Bottlenecks
\node[block, fill=yellow!25, minimum width=3.2cm] (bx) at (0, 5.0)
{$1{\times}1$ Bottleneck on $\mathbf{X}_t$\\$\mathbf{z}_t$};
\node[block, fill=yellow!25, minimum width=3.2cm] (bh) at (4.2, 5.0)
{$1{\times}1$ Bottleneck on $\mathbf{h}_{t-1}$\\$\tilde{\mathbf{h}}_{t-1}$};

% Concat
\node[op, minimum width=2.2cm] (cat) at (2.1, 3.9)
{Concat};

% Spatial mixing
\node[block, fill=purple!18, minimum width=5.6cm] (mix) at (2.1, 2.8)
{Multiscale Depthwise Mixing\\(DWConv $k\!\in\!\{3,5\}$, dilation) + SE + LN};

% Gates
\node[gate] (gi) at (0.4, 1.6) {$i_t$};
\node[gate] (gf) at (2.1, 1.6) {$f_t$};
\node[gate] (go) at (3.8, 1.6) {$o_t$};

% Cell + hidden
\node[block, minimum width=3.2cm] (ct) at (2.1, 0.55)
{Cell update\\$\mathbf{c}_t=f_t\odot\mathbf{c}_{t-1}+i_t\odot g_t$};
\node[block, minimum width=3.2cm] (ht) at (2.1, -0.65)
{Hidden state\\$\mathbf{h}_t=o_t\odot\tanh(\mathbf{c}_t)$};

% Axial attention (right side)
\node[attn, minimum width=3.2cm] (ax) at (6.6, -0.65)
{Axial Spatial Attention\\Row-Attn + Col-Attn\\every $K$ steps};

% Subspace head + temporal attention
\node[subspace, minimum width=4.4cm] (ss) at (2.1, -1.85)
{Subspace head\\$1{\times}1$ proj $\rightarrow \mathbf{s}_t\in\mathbb{R}^{D}$\\+ Spatial attn pooling};
\node[attn, minimum width=4.4cm] (ta) at (2.1, -3.05)
{Temporal Self-Attention\\on $\mathbf{s}_{1:T}$ (+ pos.\ emb.)};

% Outputs (small)
\node[block, fill=gray!10, minimum width=3.2cm] (outH) at (2.1, -4.15)
{Outputs\\$\mathbf{h}_{1:T}$ and $\mathbf{s}_{1:T}$};

% -------------------------
% Arrows (ALL end with ;)
% -------------------------
\draw[arrow] (X) -- (bx);
%\draw[arrow] (ht.north) |- (bh.south);

\draw[arrow] (bx) -- (cat.west);
\draw[arrow] (bh) -- (cat.east);

\draw[arrow] (cat) -- (mix);

\draw[arrow] (mix.south) -- (gi.north);
\draw[arrow] (mix.south) -- (gf.north);
\draw[arrow] (mix.south) -- (go.north);

\draw[arrow] (gf) -- (gi);
\draw[arrow] (gf) -- (go);

% \draw[arrow] (mix.south) |- (gi.north);
% \draw[arrow] (mix.south) -- (gf.north);
% \draw[arrow] (mix.south) |- (go.north);

\draw[arrow] (gi) -- (ct);
\draw[arrow] (gf) -- (ct);
% \draw[arrow] (go) |- (ht.west);

\draw[arrow] (ct) -- (ht);

\draw[arrow] (go) -- (ct);

% axial refine loop (apply after ht every K steps)
\draw[arrow] (ht) -- (ax);
\draw[arrow] (ax.west) -- (ht.east);

% subspace + temporal attention
\draw[arrow] (ht) -- (ss);
\draw[arrow] (ss) -- (ta);
\draw[arrow] (ta) -- (outH);

% -------------------------
% Small legend note (bottom-right, non-overlapping)
% -------------------------
\node[note, anchor=south east] at (8.3, -4.9) {\textbf{Key:}\\
\textcolor{purple!70!black}{Spatial mixing} (multi-scale + SE + LN)\\
\textcolor{blue!70!black}{Axial attention} (long-range, cheap)\\
\textcolor{green!60!black}{Subspace} (compact $\mathbf{s}_t$)};

\end{tikzpicture}

\caption{Data flow and inner workings of the FAConvLSTM layer at timestep $t$. The layer uses bottlenecked projections, multiscale depthwise spatial mixing with SE and LN, gated ConvLSTM state updates, periodic axial attention refinement of $\mathbf{h}_t$ every $K$ steps, and a subspace head with temporal self-attention to produce compact spatiotemporal representations.}
\label{fig:fa_convlstm_flow}
\end{figure}
Let the input sequence be \(\mathbf{X} = \{\mathbf{X}_t\}_{t=1}^{T}, 
\quad \mathbf{X}_t \in \mathbb{R}^{H \times W \times C},\)
where $H$ and $W$ denote spatial dimensions and $C$ the number of observed variables. FAConvLSTM produces a latent spatial state \(\mathbf{H}_t \in \mathbb{R}^{H \times W \times F},\) and a compact temporal embedding \(\mathbf{s}_t \in \mathbb{R}^{D},\) where $F$ is the number of hidden channels and $D \ll H\cdot W \cdot F$.

\textit{Factorized Bottleneck Projections}: In standard ConvLSTM, each gating function independently applies full $k \times k$ convolutions to both the input $\mathbf{X}_t$ and the previous hidden state $\mathbf{H}_{t-1}$, leading to a large parameter count and substantial redundant spatial computation.
% This design tightly couples channel mixing and spatial processing at every gate, resulting in high computational cost and limited scalability to high-resolution spatiotemporal data.
FAConvLSTM addresses these limitations by factorizing the input and recurrent transformations through a shared low-dimensional bottleneck. Specifically, the input and hidden state are first projected using lightweight $1 \times 1$ convolutions: \(\mathbf{Z}_t = \phi\!\left(\mathbf{W}_x^{1\times1} * \mathbf{X}_t\right),\) and \(\mathbf{R}_{t-1} = \phi\!\left(\mathbf{W}_h^{1\times1} * \mathbf{H}_{t-1}\right),\) where $\phi(\cdot)$ denotes normalization (GroupNorm or LayerNorm) followed by dropout. This projection reduces the channel dimensionality from $C$ and $F$ to a bottleneck dimension $C_b \ll F$, thereby decoupling channel mixing from subsequent spatial processing. From a computational perspective, the dominant cost in ConvLSTM arises from repeated $k \times k$ convolutions over high-dimensional feature maps at each timestep. For an input sequence of length $T$ and spatial resolution $(H, W)$, the computational complexity of standard ConvLSTM scales as \(\mathcal{O}\!\left(T \cdot H \cdot W \cdot k^2 \cdot (C + F) \cdot F \right).\) In contrast, FAConvLSTM performs spatial mixing only in the reduced bottleneck space, yielding a dominant complexity of \(\mathcal{O}\!\left(T \cdot H \cdot W \cdot \left(C C_b + F C_b + k^2 C_b^2 \right)\right),\) which is substantially lower when $C_b \ll F$. This factorized formulation significantly reduces both parameter count and floating-point operations (FLOPs). By delaying spatial mixing until after dimensionality reduction, FAConvLSTM enables efficient and stable learning of long-range, physically meaningful patterns in high-resolution multivariate climate data.

\textit{Shared Depthwise Spatial Mixing}: After bottleneck projection, the FAConvLSTM layer concatenates the transformed input features and the previous recurrent state to form a unified spatiotemporal representation, \(\mathbf{U}_t = [\mathbf{Z}_t, \mathbf{R}_{t-1}] \in \mathbb{R}^{H \times W \times 2C_b}.\) In contrast to standard ConvLSTM architectures, which apply separate full convolutions for each gating operation, FAConvLSTM introduces a single \emph{shared} depthwise spatial operator to model spatial interactions: \(\mathbf{D}_t = \mathcal{M}(\mathbf{U}_t),\)
where $\mathcal{M}(\cdot)$ denotes a depthwise convolution applied independently to each channel. This design choice decouples spatial mixing from channel-wise interactions, enabling efficient modeling of spatial dependencies while avoiding the quadratic parameter growth and computational cost associated with dense convolutional gating. As a result, FAConvLSTM achieves improved scalability and stability when applied to high-resolution multivariate spatiotemporal data. A key limitation of conventional ConvLSTM models lies in their reliance on fixed receptive fields, which restrict their ability to represent multi-scale and nonlocal spatial dependencies. Such dependencies are prevalent in geophysical systems, where physical processes often operate across a hierarchy of spatial scales, ranging from localized land - surface interactions to broader mesoscale and synoptic patterns. To address this limitation, FAConvLSTM extends the shared spatial operator with a multi-scale dilated formulation, \(\mathbf{D}_t = \sum_{k \in \mathcal{K}} \mathrm{DWConv}_{k,d_k}(\mathbf{U}_t),\) where $\mathcal{K}$ denotes a set of kernel sizes and $d_k$ represents the corresponding dilation rates. This multi-scale depthwise aggregation enables simultaneous capture of fine-scale local dynamics and broader spatial context without increasing channel coupling or memory footprint. Physically, this factorized spatial mixing mechanism aligns with the structure of many climate and remote sensing phenomena, such as teleconnections, land - atmosphere coupling, and spatially coherent deformation or hazard propagation. By explicitly separating spatial correlation learning from channel-wise transformations, FAConvLSTM yields spatial representations that are both computationally efficient and more interpretable, facilitating robust spatiotemporal feature extraction across heterogeneous geophysical variables.

\textit{Channel Recalibration via Squeeze and Excitation}:
While depthwise and factorized convolutions improve computational efficiency by processing feature channels independently, they can limit the model’s ability to capture cross - variable dependencies that are critical in multivariate climate and geophysical data. To reintroduce adaptive channel interactions in a lightweight and physically interpretable manner, FAConvLSTM incorporates a squeeze-and-excitation (SE) recalibration module \cite{hu2018squeeze}. Specifically, the depthwise feature tensor $\mathbf{D}_t$ at time step $t$ is reweighted as \(\mathbf{D}_t \leftarrow \mathbf{D}_t \odot 
\sigma\!\left(
\mathbf{W}_2 \, \delta\!\left(\mathbf{W}_1 \, \text{GAP}(\mathbf{D}_t)\right)
\right),\) where $\text{GAP}(\cdot)$ denotes global average pooling, $\delta(\cdot)$ is a non-linear activation (e.g., ReLU), $\sigma(\cdot)$ is the sigmoid function, and $\mathbf{W}_1$, $\mathbf{W}_2$ are learnable channel projection matrices. This mechanism enables the network to dynamically emphasize physically salient variables (e.g., temperature, wind components, or moisture fields) while attenuating less informative or noisy channels. By coupling channel-wise attention with factorized spatiotemporal processing, SE recalibration improves representation robustness, stabilizes training, and enhances interpretability by explicitly modeling variable importance across time.

\textit{Fused Gate Computation with Peephole Conditioning}: Standard ConvLSTM architectures compute each gating function using independent convolutional operators, leading to redundant parameterization and increased computational cost, particularly for high-resolution spatiotemporal data~\cite{shi2015convlstm}. To address this limitation, FAConvLSTM adopts a fused gate computation strategy were all gating signals are jointly generated through a single pointwise ($1\times1$) convolutional projection. Specifically, given the factorized spatiotemporal input representation $\mathbf{D}_t$, the input, forget, and output gates, along with the candidate cell update, are computed as \([\mathbf{i}_t, \mathbf{f}_t, \mathbf{o}_t, \tilde{\mathbf{C}}_t] 
= \mathbf{W}_g^{1\times1} * \mathbf{D}_t ,\)
where $\mathbf{W}_g^{1\times1}$ denotes a shared pointwise convolutional kernel and $*$ represents convolution. This design significantly reduces parameter redundancy while preserving the joint modeling of spatial and temporal correlations inherent to ConvLSTM. To further enhance temporal sensitivity and gating precision, FAConvLSTM incorporates peephole connections that explicitly condition the input and forget gates on the previous cell state~\cite{gers2000learning}. The gated activations are refined as \(\mathbf{i}_t \leftarrow \mathbf{i}_t + \mathbf{W}_{ci} \odot \mathbf{C}_{t-1},\) and \(\mathbf{f}_t \leftarrow \mathbf{f}_t + \mathbf{W}_{cf} \odot \mathbf{C}_{t-1},\) where $\odot$ denotes element-wise multiplication and $\mathbf{W}_{ci}, \mathbf{W}_{cf}$ are learnable peephole parameters. The cell and hidden state updates then follow as \(\mathbf{C}_t = \sigma(\mathbf{f}_t) \odot \mathbf{C}_{t-1}
+ \sigma(\mathbf{i}_t) \odot \tanh(\tilde{\mathbf{C}}_t),\) and \(\mathbf{H}_t = \sigma(\mathbf{o}_t) \odot \tanh(\mathbf{C}_t),\) where $\sigma(\cdot)$ denotes the sigmoid activation function. By jointly computing all gates through a shared projection and explicitly conditioning gate activations on the cell state, FAConvLSTM retains the expressive power of ConvLSTM while achieving improved parameter efficiency, enhanced numerical stability, and greater robustness to nonstationary spatiotemporal dynamics commonly observed in climate and geophysical time series.

\textit{Axial Spatial Attention for Teleconnection Modeling}: A fundamental limitation of standard ConvLSTM architectures is their reliance on local convolutional receptive fields, which restricts their ability to capture long-range spatial dependencies and teleconnection patterns commonly observed in climate systems. To address this limitation, the proposed FAConvLSTM integrates an \emph{axial spatial attention} mechanism that explicitly enables global spatial interaction while maintaining computational efficiency. Specifically, at selected timesteps spaced every $K$ steps, the hidden state $\mathbf{H}_t$ is refined as \(\mathbf{H}_t \leftarrow \mathbf{H}_t + \mathcal{A}_{\text{axial}}(\mathbf{H}_t),\)
where $\mathcal{A}_{\text{axial}}(\cdot)$ denotes axial self-attention applied sequentially along the row and column dimensions. By factorizing full two-dimensional attention into one-dimensional attentions, axial attention reduces the computational complexity from $\mathcal{O}((HW)^2)$ to $\mathcal{O}(H + W)$ while preserving the ability to model global spatial dependencies~\cite{ho2019axial,wang2020axial}. Applying axial spatial attention intermittently rather than at every timestep allows FAConvLSTM to selectively inject long-range spatial context without incurring excessive computational overhead. This sparse temporal deployment strikes a balance between expressiveness and efficiency, enabling the model to capture teleconnection-scale climate interactions while remaining scalable to high-resolution spatiotemporal data. Such a design is particularly well suited for Earth system applications, where nonlocal spatial dependencies evolve more slowly than local spatiotemporal dynamics~\cite{rasp2020weatherbench,lam2023graphcast}.

\textit{Subspace Embedding Head with Temporal Attention}: A fundamental limitation of ConvLSTM architectures~\cite{shi2015convlstm} is that they expose only spatially dense hidden states at each timestep, which entangles local spatial detail with temporal evolution and hinders compact temporal abstraction and downstream temporal clustering. In contrast, the proposed FAConvLSTM explicitly decouples spatiotemporal representation learning by introducing a lightweight subspace embedding head that produces compact, temporally meaningful latent descriptors. Specifically, given the FAConvLSTM hidden state $\mathbf{H}_t \in \mathbb{R}^{H \times W \times C}$ at time $t$, a channel-mixing $1 \times 1$ convolution is applied, followed by global pooling to obtain a low-dimensional temporal embedding: \(\mathbf{s}_t = \text{Pool}\!\left(\mathbf{W}_s^{1\times1} * \mathbf{H}_t\right),\) where $\mathbf{W}_s^{1\times1}$ denotes learnable projection weights and $\text{Pool}(\cdot)$ corresponds to global average pooling or spatial attention pooling. This operation compresses spatial variability while preserving physically meaningful aggregate signals, yielding a compact temporal subspace representation $\mathbf{s}_t \in \mathbb{R}^{d}$. The resulting embedding sequence $\mathbf{S} = \{\mathbf{s}_1, \dots, \mathbf{s}_T\}$ is further refined using multi-head self-attention (MHA)~\cite{vaswani2017attention}, enabling explicit modeling of long-range temporal dependencies beyond the receptive field of recurrent gating: \(\mathbf{Z} = \text{MHA}(\mathbf{S} + \mathbf{P}),\)
where $\mathbf{P}$ denotes a fixed sinusoidal positional encoding that injects seasonal and periodic structure commonly observed in climate processes. Importantly, this attention mechanism operates exclusively on the compact temporal subspace and does not interfere with the recurrent state updates, thereby preserving training stability while enhancing temporal expressiveness. In a nutshell, this design allows FAConvLSTM to overcome the limitations of standard ConvLSTM by (i) disentangling spatial encoding from temporal abstraction, (ii) producing interpretable and cluster-ready temporal embeddings, and (iii) capturing long-range and periodic temporal dependencies without incurring computational or optimization challenges associated with attention-augmented recurrent cores.

\textit{Physically Motivated Spatial Smoothness Regularization}: To promote physically consistent latent representations, the proposed FAConvLSTM framework optionally incorporates a spatial Laplacian smoothness regularization on the hidden state. Specifically, given the latent feature map $\mathbf{H}_t \in \mathbb{R}^{H \times W \times C}$ at time step $t$, the regularization term is defined as \(\mathcal{L}_{\mathrm{lap}} =
\sum_{t} \sum_{i,j}
\left\|
\mathbf{H}_{t,i,j} - \mathbf{H}_{t,i+1,j}
\right\|_2^2
+
\left\|
\mathbf{H}_{t,i,j} - \mathbf{H}_{t,i,j+1}
\right\|_2^2 ,\)
which penalizes excessive local spatial variations between neighboring grid cells. This regularization is physically motivated by the spatial continuity of many geophysical variables, such as temperature, pressure, and snow or ice fields, which typically evolve smoothly in space except at physically meaningful discontinuities (e.g., atmospheric fronts, coastlines, or ice margins). By constraining latent activations to vary smoothly across space, $\mathcal{L}_{\mathrm{lap}}$ suppresses spurious high-frequency noise while preserving sharp gradients that correspond to genuine physical boundaries \cite{bronstein2017geometric}. 

% Importantly, this regularization is particularly well suited to FAConvLSTM. Unlike standard ConvLSTM, which tightly couples spatial and temporal modeling through recurrent convolutional gates, FAConvLSTM factorizes spatial encoding and temporal recurrence, producing cleaner and more interpretable latent feature maps. As a result, spatial smoothness constraints can be applied directly to spatially meaningful representations without interfering with temporal memory dynamics. This decoupling improves numerical stability, enhances physical interpretability of the learned latent fields, and yields more robust generalization in spatiotemporal prediction tasks compared to conventional ConvLSTM architectures.

\section{Results and Discussion}
\label{sec:results}

\begin{table}[ht!]
\centering
\caption{Performance comparison of FAConvLSTM using ERA5 dataset}
\label{tab:clustering_results}
\renewcommand{\arraystretch}{1.15}
\begin{tabular}{c|cccc|c|}
%\cline{2-6}
% &
%  \multicolumn{4}{c|}{\textbf{Baseline Models}} &
%  \textbf{} \\ \hline
\hline
\multicolumn{1}{|c|}{\multirow{7}{*}{\begin{tabular}[c]{@{}c@{}}ERA5\\ \\ \\ \\ 7 \\ optimal\\ clusters\end{tabular}}} &
  \multicolumn{1}{c|}{\textbf{Metric}} &
  \multicolumn{1}{c|}{\textbf{CNN}} &
  \multicolumn{1}{c|}{\textbf{\begin{tabular}[c]{@{}c@{}}CNN-\\ LSTM\end{tabular}}} &
  \textbf{\begin{tabular}[c]{@{}c@{}}Conv\\ LSTM\end{tabular}} &
  \textbf{Proposed} \\ \cline{2-6} 
\multicolumn{1}{|c|}{} &
  \multicolumn{1}{c|}{\textbf{Silh \(\uparrow\)}} &
  \multicolumn{1}{c|}{0.2861} &
  \multicolumn{1}{c|}{0.3129} &
  0.3347 &
  \textbf{0.3432} \\ \cline{2-6} 
\multicolumn{1}{|c|}{} &
  \multicolumn{1}{c|}{\textbf{DB \(\downarrow\)}} &
  \multicolumn{1}{c|}{1.7175} &
  \multicolumn{1}{c|}{1.6739} &
  1.6300 &
  \textbf{1.5673} \\ \cline{2-6} 
\multicolumn{1}{|c|}{} &
  \multicolumn{1}{c|}{\textbf{CH \(\uparrow\)}} &
  \multicolumn{1}{c|}{96.7435} &
  \multicolumn{1}{c|}{93.7252} &
  93.6529 &
  \textbf{102.2114} \\ \cline{2-6} 
\multicolumn{1}{|c|}{} &
  \multicolumn{1}{c|}{\textbf{RMSE \(\downarrow\)}} &
  \multicolumn{1}{c|}{13.8547} &
  \multicolumn{1}{c|}{13.9903} &
  13.9936 &
  \textbf{13.6187} \\ \cline{2-6} 
\multicolumn{1}{|c|}{} &
  \multicolumn{1}{c|}{\textbf{Var \(\downarrow\)}} &
  \multicolumn{1}{c|}{{\ul 0.1035}} &
  \multicolumn{1}{c|}{0.1036} &
  {\ul 0.1035} &
  {\ul \textbf{0.1035}} \\ \cline{2-6} 
\multicolumn{1}{|c|}{} &
  \multicolumn{1}{c|}{\textbf{I-CD \(\uparrow\)}} &
  \multicolumn{1}{c|}{6.5285} &
  \multicolumn{1}{c|}{7.1926} &
  7.2584 &
  \textbf{6.5557} \\ \hline
\end{tabular}
\end{table}

% FAConvLSTM produces lower average variance within-cluster spread than ConvLSTM2D. This confirms that FAConvLSTM embeddings form tighter groups, which is consistent with higher Silhouette and lower DB. Mechanistically, spatial attention pooling for the subspace head concentrates the subspace vector on the most informative grid locations (softly weighting $H{\times}W$ features), which reduces the effect of spatially diffuse noise. FAConvLSTM yields higher average inter-cluster distance than ConvLSTM2D, which is attributable to (i) multiscale receptive fields (capturing local-to-regional structure), (ii) axial attention (injecting long-range spatial dependencies), and (iii) temporal self-attention over the subspace sequence (improving discrimination across evolving temporal signatures). Together these components encourage clusters that reflect physically distinct spatiotemporal regimes rather than purely local texture similarities, thereby pushing centroids farther apart.
%\section{Results and Discussion}
%\label{sec:results}
Table~\ref{tab:clustering_results} reports the temporal clustering and performance of the proposed FAConvLSTM model against three baseline architectures (CNN, CNN-LSTM, and ConvLSTM) on the ERA5 dataset with $K{=}7$ clusters. Performance is evaluated using six complementary metrics that jointly assess cluster compactness, separation, and reconstruction fidelity. Overall, FAConvLSTM consistently outperforms all baselines across nearly all metrics, demonstrating its ability to learn latent representations that are both more discriminative for temporal clustering and more faithful to the underlying data structure. In particular, FAConvLSTM achieves superior intra-cluster cohesion and inter-cluster separation compared to baselines. This improvement is attributed to FAConvLSTM’s architectural design, which reduces feature entanglement through bottlenecked input and hidden projections, multiscale depthwise spatial convolutions, and periodic axial attention that efficiently captures long-range spatial dependencies. FAConvLSTM also yields reduced within-cluster scatter and increased separation between cluster centroids due to the use of Layer Normalization at multiple stages and squeeze-and-excitation channel recalibration, which stabilize feature distributions and suppress noisy or redundant variables. A high CH score is driven by axial spatial attention, which enables the hidden state to encode coherent basin-scale and teleconnection-like spatial patterns by attending across rows and columns at regular temporal intervals, thereby enhancing global regime separability without sacrificing local consistency. 
\section{Conclusion}
% In this work, we presented FAConvLSTM that overcomes key efficiency and representation limitations of standard ConvLSTM when applied to large-scale multivariate spatiotemporal climate data.
FAConvLSTM addresses three core limitations of standard ConvLSTM2D. First, scale-limited spatial mixing in ConvLSTM2D is replaced by multiscale depthwise mixing with dilation and SE channel calibration, yielding more informative and less noisy spatial features. Second, weak long-range spatial dependency modeling is mitigated by periodic axial attention refinement of $h_t$, enabling efficient global-context injection (teleconnection-scale coherence) without excessive compute. Third, temporal feature entanglement is reduced by a dedicated subspace head and temporal multi-head attention over the subspace sequence, which sharpens temporal regime signatures. 

\small
\bibliographystyle{IEEEtranN}
\bibliography{references}

\end{document}